\documentclass{article}

\usepackage{microtype}
\usepackage{graphicx}
\usepackage{subfigure}
\usepackage{booktabs} 
\usepackage{tgbonum}
\usepackage[T1]{fontenc}

\usepackage{hyperref}

\usepackage[accepted]{icml2024}

\usepackage{amsmath}
\usepackage{amssymb}
\usepackage{mathtools}
\usepackage{amsthm}

\usepackage{fancyvrb}
\usepackage{tcolorbox}
\usepackage{spverbatim}
\usepackage{listings}
\usepackage[capitalize,noabbrev]{cleveref}

\theoremstyle{plain}

\theoremstyle{definition}

\theoremstyle{remark}

\usepackage[textsize=tiny]{todonotes}

\usepackage{multirow}

\usepackage{tabularx}

\icmltitlerunning{In Search of Needles in a 11M Haystack}

\begin{document}

\twocolumn[
\icmltitle{In Search of Needles in a 11M Haystack: \\ Recurrent Memory Finds What LLMs Miss}



\icmlsetsymbol{equal}{*}

\begin{icmlauthorlist}
\icmlauthor{Yuri Kuratov}{airi,dp}
\icmlauthor{Aydar Bulatov}{dp}
\icmlauthor{Petr Anokhin}{airi}
\icmlauthor{Dmitry Sorokin}{airi}
\icmlauthor{Artyom Sorokin}{airi}
\icmlauthor{Mikhail Burtsev}{lims}
\end{icmlauthorlist}

\icmlaffiliation{airi}{AIRI, Moscow, Russia}
\icmlaffiliation{dp}{Neural Networks and Deep Learning Lab, MIPT, Dolgoprudny, Russia}
\icmlaffiliation{lims}{London Institute for Mathematical Sciences, London, UK}

\icmlcorrespondingauthor{Yuri Kuratov}{yurii.kuratov@phystech.edu}
\icmlcorrespondingauthor{Aydar Bulatov}{bulatov.as@phystech.edu}
\icmlcorrespondingauthor{Mikhail Burtsev}{mb@lims.ac.uk}


\vskip 0.3in
]



\printAffiliationsAndNotice{}  

\begin{abstract}
This paper addresses the challenge of processing long documents using generative transformer models. To evaluate different approaches, we introduce BABILong, a new benchmark designed to assess model capabilities in extracting and processing distributed facts within extensive texts. Our evaluation, which includes benchmarks for GPT-4 and RAG, reveals that common methods are effective only for sequences up to $10^4$ elements. In contrast, fine-tuning GPT-2 with recurrent memory augmentations enables it to handle tasks involving up to $11\times 10^6$ elements. This achievement marks a substantial leap, as it is by far the longest input processed by any neural network model to date, demonstrating a significant improvement in the processing capabilities for long sequences.
\end{abstract}

\section{Introduction}
\label{intro}

Memory is an essential component of both natural and artificial cognitive systems. Different types of memory represent general knowledge about some domain, facts, and episodic or specific information. Training of neural models encodes representations for general knowledge and repeating facts in network parameters. On the other hand, task-specific information is provided as a part of the model's input. This type of information is particularly important because it conditions the generation of a solution.

Recent progress in machine learning has resulted in the extension of input size for commonly used models by three orders of magnitude, from hundreds to hundreds of thousands of elements. However, further increase in input sequence length is limited by the quadratic scaling of compute required for the calculation of self-attention in transformers. Today, new models are emerging that challenge these limitations and have the potential to go beyond millions of input elements.

\begin{figure}[t]
\centering
\includegraphics[width=\linewidth]{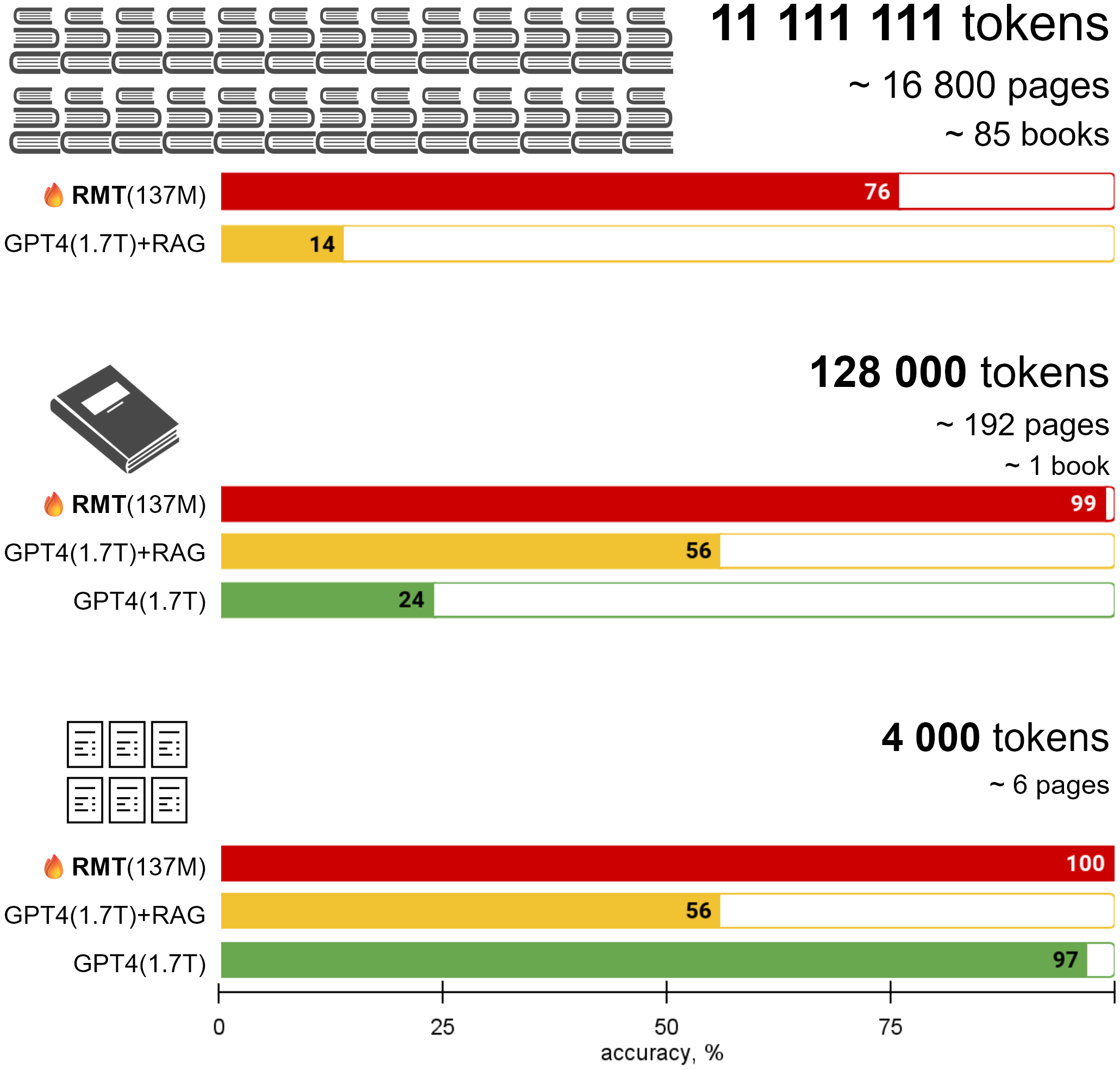}
\vspace{-6mm}
\caption{\textbf{Memory augmented transformer answers questions about facts hidden in very long texts when retrieval augmented generation fails.} We create a new BABILong dataset by randomly distributing simple episodic facts inside book corpus. Common RAG method fails to answer questions because order of facts matters. GPT-4 LLM effectively uses only fraction of the context and falls short for a full 128K window. Small LM (GPT-2) augmented with recurrent memory and fine-tuned for the task generalise well up to record 11M tokens. The parameter count for GPT-4 is based on public discussions.}
\vspace{-3mm}
\label{fig:summary}
\end{figure}

One of the promising approaches to extend the context window of transformers is augmentation with recurrent memory \cite{rmt_2022}. Here, a pre-trained language model is extended with internal recurrent memory and fine-tuned to use this memory to solve specific tasks for long contexts split into segments. For this class of models, computational complexity scales linearly with input size.

In this work, we further develop the recurrent memory approach by adding in-context retrieval based on the recurrent memory embedding of input segments. As the field evolves rapidly, methods for evaluating models with extremely long inputs lag behind. Recent benchmarks for understanding large contexts, such as LongBench \cite{bai2023longbench}, include tasks with lengths only up to $4\cdot10^4$. To test our models with much longer contexts, we build upon the first 'needle in a haystack' tests and propose the new BABILong framework to 'hide' algorithmically generated question answering and reasoning problems inside a corpus of book texts.

The main contributions of our work are as follows:

1. The introduction of BABILong, a novel generative benchmark for evaluating the performance of NLP models in processing arbitrarily long documents with distributed facts.

2. Evaluation of GPT-4 and RAG on 'needle in a haystack' question answering tasks up to millions of tokens.

3. Evaluation of recurrent memory transformer on input texts up to 11 million tokens sets a new record for the sequence size processed by a single model, extending the known capabilities of neural networks.

The BABILong benchmark and examples with LLMs evaluation are available: \url{https://github.com/booydar/babilong}.

\section{BABILong: Needle in a haystack benchmark for long document processing}
\label{dataset}

\begin{figure}[t]
\centering
\includegraphics[width=.75\linewidth]{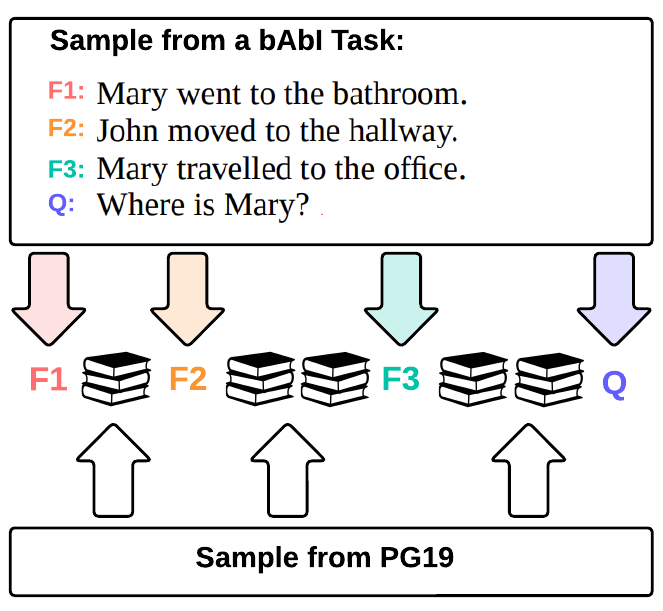}
\caption{\textbf{Example generation for BABILong dataset.} Statements relevant for the question from a bAbILong sample are hidden inside a larger irrelevant texts from PG19.}
\label{babilong_sample_generation}
\end{figure}


The fundamental concept behind the \textbf{B}enchmark for \textbf{A}rtificial \textbf{I}ntelligence for \textbf{Long}-context evaluation is to extend the length of existing tasks to evaluate the ability of generative models to efficiently handle long contexts. Solving tasks with a long context size requires the model to distinguish important information from large amounts of irrelevant details. To simulate this behavior we "hide" the sentences of the original task between the sentences of irrelevant text that is drawn from another closely related distribution (see Figure~\ref{babilong_sample_generation}). Examples are constructed by gradually adding new sentences from the background dataset in their natural order until the augmented sample reaches the desired length. This way, we are not bound by the length of the original task itself, making it possible to assess even the longest available models with context sizes up to millions of tokens.
For background text we use books from the PG19 dataset ~\citep{rae2019compressive} due to the substantial book lengths and naturally occurring long contexts. The model is required first to distinguish the sentences related to the original task, then memorize and subsequently utilize them to generate the correct solution.

In this work we focus on extending the bAbI benchmark~\cite{WestonBCM15}, that consists of 20 tasks designed for evaluation of basic aspects of reasoning. The bAbI tasks are generated by simulating a set of characters and objects engaged in various movements and interactions with each other in multiple locations. Each interaction is represented by a fact, e.g. "Mary travelled to the office", and the task is to answer a question using the facts from the current simulation, for instance, "Where is Mary?". The bAbI tasks vary based on the number of facts, question complexity and the aspects of reasoning they evaluate, namely spatial and temporal reasoning, deduction, coreference resolution and more. In the paper we label tasks from 'qa1' to 'qa20'. First ten tasks are listed in Table~\ref{tab:task_eda} and examples of samples produced following our pipeline can be found in the Appendix~\ref{appendix:babi_samples}.

\begin{table}[h]
\caption{\small First ten tasks of BABILong  with number of supporting and distracting facts.}
\label{tab:task_eda}
\begin{center}
\begin{small}
\begin{sc}
\fontsize{7}{8}
\selectfont 
\begin{tabular}{llcc}
\toprule
Task          & Name & min facts  & max facts  \\
          &  & per task &  per task \\
\midrule
qa1        &  single supporting fact         &  2          & 10 \\
qa2        &  two supporting facts           &  2          & 68 \\
qa3        &  three supporting facts         &  4          & 320 \\
qa4        &  two arg relations              &  2          & 2  \\
qa5        &  three arg relations            &  2          & 126 \\
qa6        &  yes-no questions               &  2          & 26 \\
qa7        &  counting                       &  2          & 52 \\
qa8        &  lists-sets                     &  2          & 50 \\
qa9        &  simple negation                &  2          & 10 \\
qa10       &  indefinite knowledge           &  2          & 10 \\
\bottomrule
\end{tabular}
\end{sc}
\end{small}
\end{center}
\end{table}

Most NLP benchmarks are vulnerable to data leakage to enormous training sets of modern large language models~\cite{sainz-etal-2023-nlp}. Generated benchmarks, such as bAbI and BABILong, are not prone to this type of contamination.

It is important to emphasise that in this work we intentionally employ simple algorithmic tasks to highlight the fundamental limitations of current models in collecting evidence over long contexts even for a simple reasoning. Nevertheless, we posit that the 'needle in a haystack' approach can be applied to incorporate more complex tasks, using the same strategy of mixing task sentences with background text.

\section{Transformers with in context and vector based retrieval on BABILong}

State-of-the-art large language models have an impressive size of the context window. However, due to the additive nature of self-attention, the growing length of the distracting text should affect the performance of the model, which tries to locate a few supporting facts. For our experiments, we selected GPT-4-Turbo \cite{Achiam2023GPT4TR} with a context window of  128k tokens and Mistral \cite{Jiang2023Mistral7} with a context length of 32k tokens. We performed our benchmarks for GPT-4-Turbo on 'qa1'-'qa5' tasks and for Mistral on 'qa1'-'qa10' tasks. In each task, we exponentially grow the length of the input size. Prompts used to evaluate the models are listed in Appendix \ref{sec:prompt}. 

\begin{figure}
\centering
\includegraphics[width=0.7\linewidth]{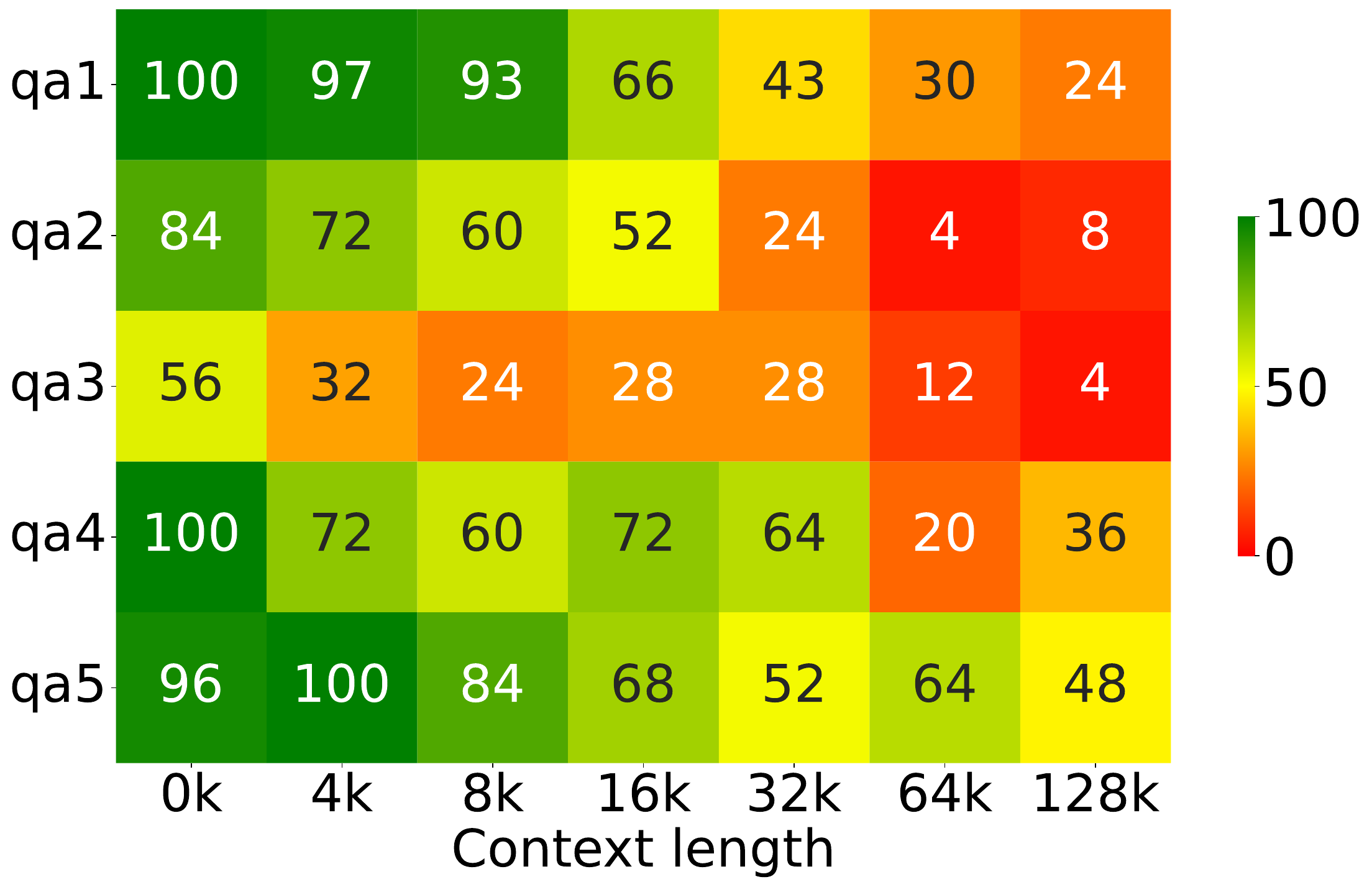}
\caption{\textbf{GPT-4 fails to solve needle in a haystack tasks for 75 \% of available context window.} Every row shows accuracy in \% of solving corresponding BABILong task ('qa1'-'qa5') and every column corresponds to the task size submitted to GPT-4-Turbo with 128K context window. All values are averages of 25 samples.}
\label{fig:gpt_eval}
\end{figure}

We present evaluation results for the GPT-4-Turbo in Fig.~\ref{fig:gpt_eval}. It is seen that without context (0k) GPT-4 can solve tasks 'qa1', 'qa4', and 'qa5' with almost 100\% accuracy. The most difficult task for GPT-4 is 'qa3' which requires complex reasoning about past locations of persons and objects. In all test tasks, when we grow the context length up to 128k tokens the accuracy of GPT-4 decreases. We also studied how the performance of the GPT-4-Turbo is affected by the positions of the facts in the input query. Refer to the Appendix \ref{sec:llmloc} for details.

\begin{figure}[t]
\centering
\includegraphics[width=0.63\linewidth]{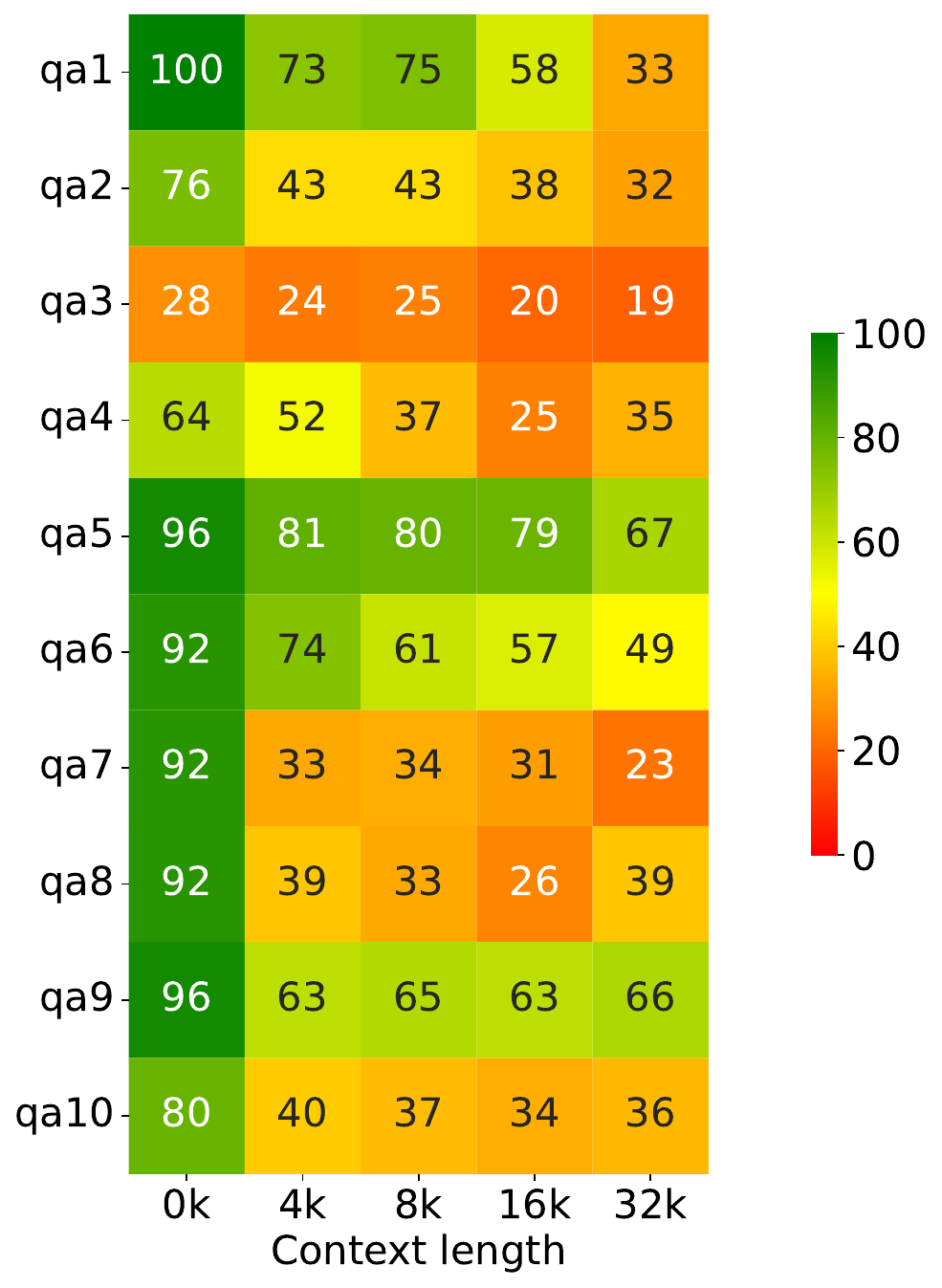}
\caption{\textbf{Mistral performance scales only for some tasks but quickly degenerates for majority of others as context grows.} Every row shows accuracy in \% of solving corresponding BABILong task ('qa1'-'qa10') and every column corresponds to the task size submitted to Mistarl-Medium with 32K context window. All values are averages of 25 samples.}
\label{fig:mistral_eval}
\end{figure}

Similar results were observed for the Mistral model (Fig.~\ref{fig:mistral_eval}). It is seen, that in general GPT-4 with the same prompt achieves higher accuracy than the Mistral, but both models start failing when the context length increases. This highlights that even models with the largest context windows can't efficiently identify facts when the amount of distracting text is extremely high.

\begin{figure}[t]
\centering
\includegraphics[width=0.6\linewidth]{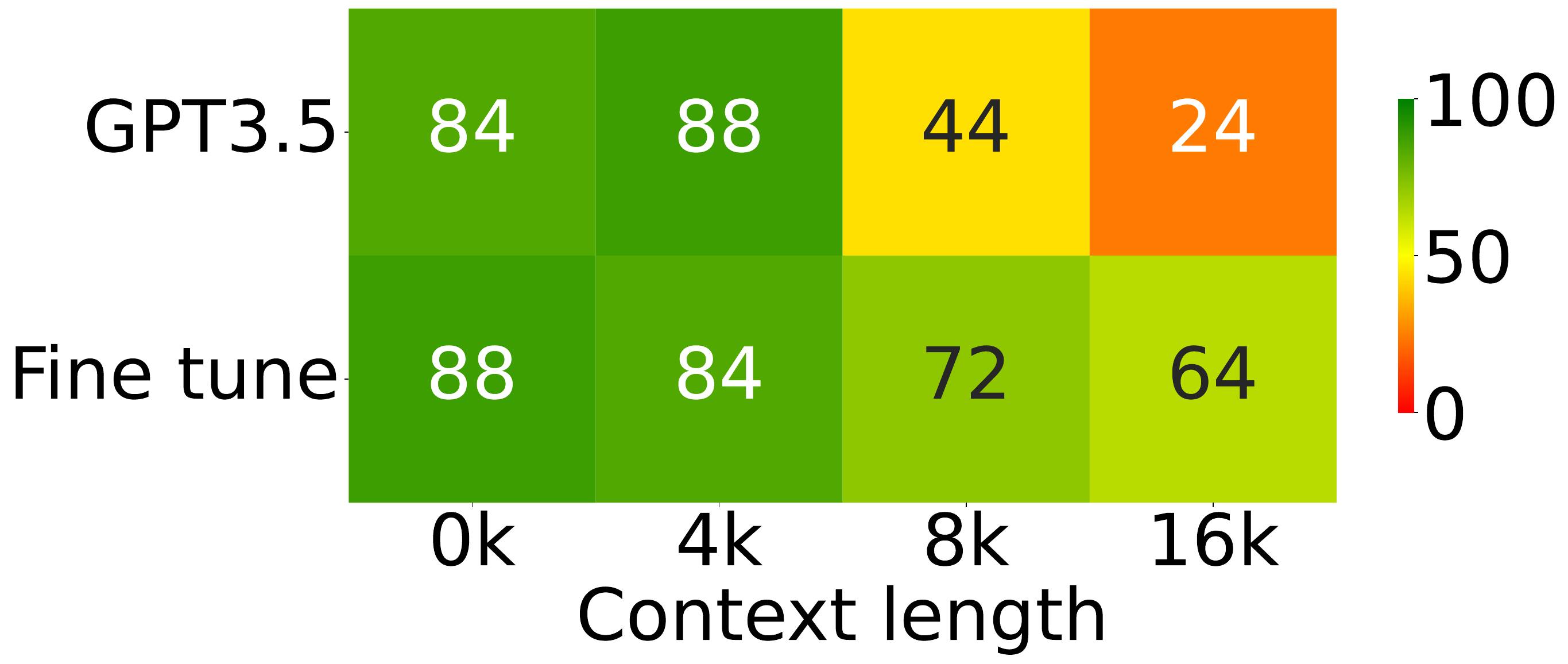}
\caption{\textbf{Fine-tuning of GPT-3.5 improves search of supporting facts in medium context size.} Every row shows accuracy in \% for GPT-3.5 before and after fine-tuning via API with 100 samples. Every column corresponds to the task size. All values are averages of 25 samples.}
\label{fig:finetune}
\end{figure}

OpenAI provides a service for fine-tuning GPT-3.5 models with custom data. We evaluated GPT-3.5 model fine-tuned with 100 'qa1' tasks for 3 epochs. The length of each task was 16k tokens. The evaluation results are shown in Fig.~\ref{fig:finetune}. It is seen that fine-tuning improves the quality of GPT-3.5 at the largest possible context equal to 16k tokens. However, the performance of the model fine-tuned with 100 examples still decreases when we increase the amount of noise. 

In this study, we employed the FAISS \cite{douze2024faiss} vector database, using Langchain library \cite{Chase_LangChain_2022}, for our experimental RAG setup. We utilized the 'text-embedding-ada-002' model for generating text embeddings. Our methodology encompassed two distinct approaches for text chunking: firstly, segmentation by sentences utilizing the NLTK library, and secondly, division into segments of 512 tokens each. We adopted a binary metric for evaluating retrieval accuracy, where the criterion was the presence or absence of relevant facts (singular or multiple, based on the specific task) within the retrieved text chunks. This retrieval accuracy was quantified for the top 5 chunks. Additionally, we assessed the performance of GPT-4-turbo in conjunction with the retrieved facts, specifically focusing on the 'qa1' task. Our experimental scope spanned various context lengths, including 8k, 64k, and 128k tokens for tasks 'qa1' through 'qa5' of the BABILong dataset, with added 4k, 16k, 32k, 500k, 1M and 10M token length for an in-depth analysis of the 'qa1' task. Additionally, we assessed the performance of RAG on the 'qa1' task, utilizing precomputed Wikipedia embeddings\footnote{\url{https://huggingface.co/datasets/Supabase/wikipedia-en-embeddings}} instead of pg-19  with an average embedding size of 250 tokens. This evaluation aimed to determine the influence of embedding size and noise characteristics on model performance. For each task, we maintained a consistent sample size of 50 across different context lengths.

\begin{figure}[t]
\centering
\includegraphics[width=1.0\linewidth]{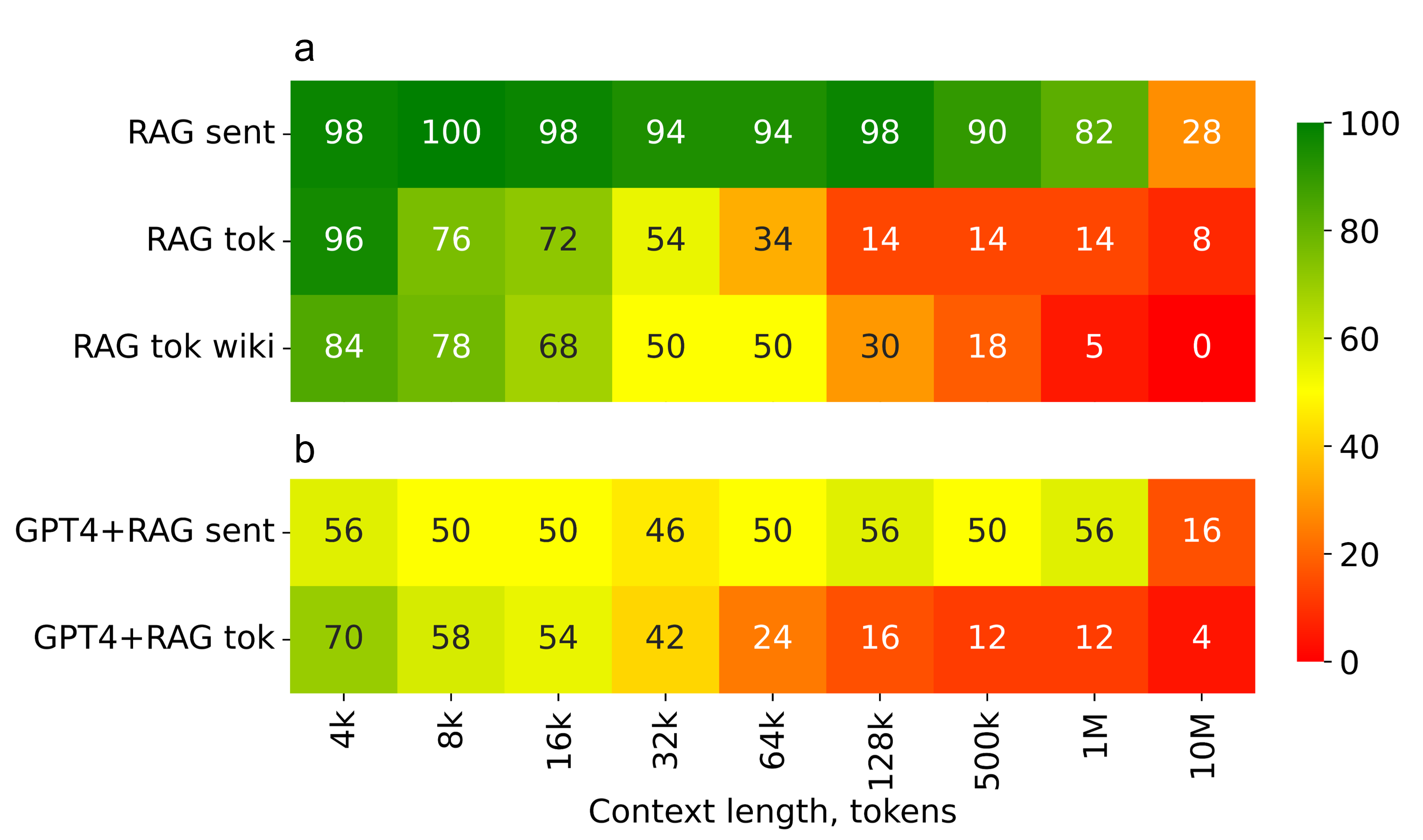}
\caption{\textbf{Retrieval augmentation does not help to solve needle in a haystack QA task.} (a) Top 5 recall scores of a retrieval RAG component 'qa1' task for the given size for sentences (sent) and text pieces of 512 tokens (tok).  
(b) Accuracy in \% by GPT-4 based RAG. All values are averages of 50 samples. }
\label{fig:rag_qa1_full}
\end{figure}

\begin{figure}[t]
\centering
\includegraphics[width=0.8\linewidth]{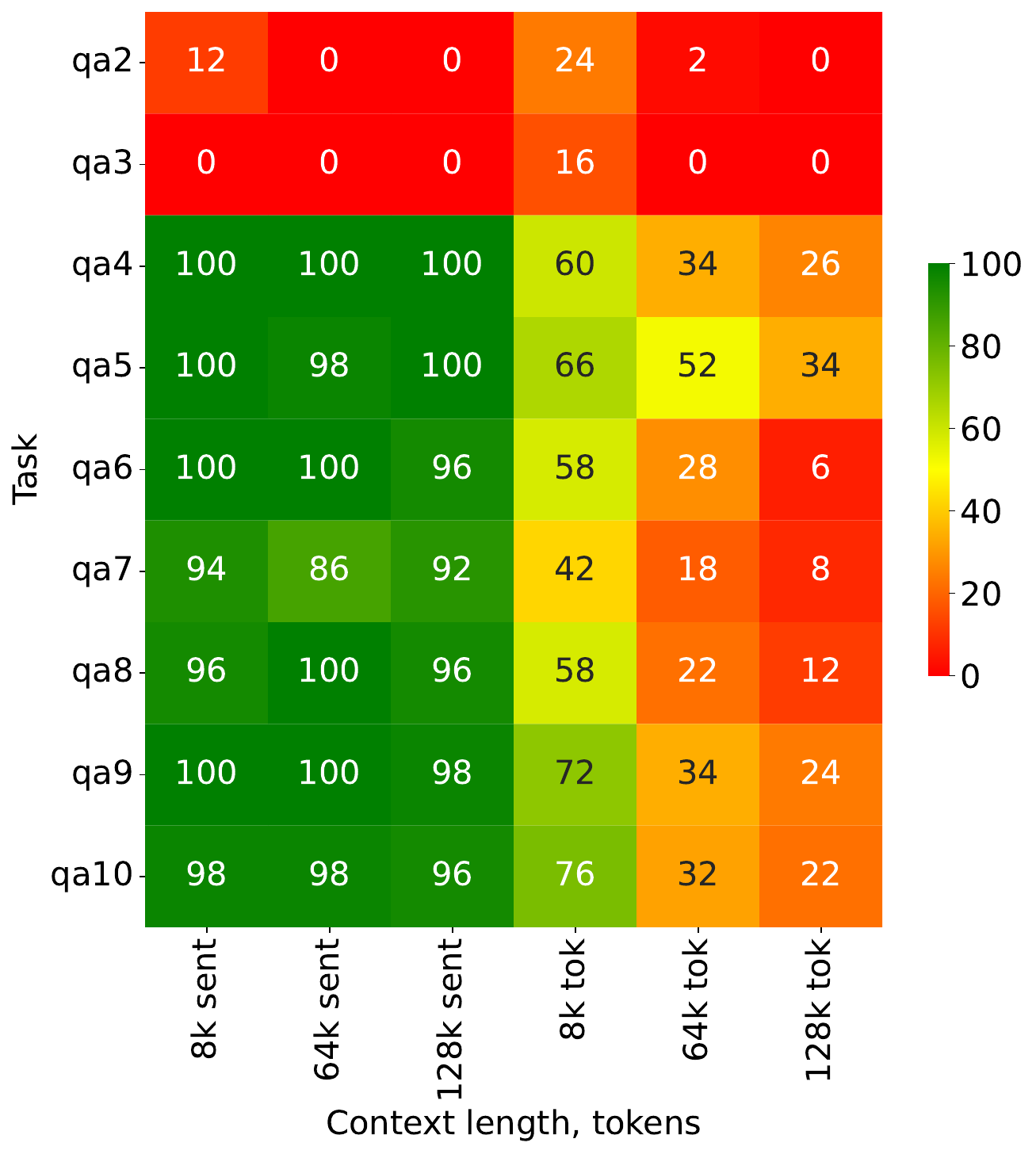}
\caption{\textbf{Retrieval performance critically depends on a task and amount of text per embedding.}  Evaluation results for vector-based retrieval on 'qa2'-'qa10' tasks.}
\label{fig:rag_qa2_10}
\end{figure}

The findings from the 'qa1' task, depicted in the Fig.~\ref{fig:rag_qa1_full}a, indicate that retrieval performance using sentence chunks was superior to that of 512-token segments, with a notable decrease in accuracy observed only at the 10M token context length. However, this superiority is task-specific and may not translate effectively to real-world applications due to the potential for information loss in smaller chunks.

Full RAG pipeline with GPT-4-turbo shows mediocre but scalable performance for sentence embeddings and poor scalability with chunk embeddings as shown on the Fig.~\ref{fig:rag_qa1_full}b. Weak performance of RAG might be attributable to the temporal dependencies inherent in the task, wherein the relevant fact is positioned at the end of the text. The default retrieval algorithm's lack of temporal consideration underscores the necessity for manually incorporating this aspect in tasks sensitive to temporal dynamics.

The sentence chunking method demonstrated superior retrieval performance compared to the 512-token segmentation at tasks 'qa4' - 'qa10', yet both methods were unsuccessful in tasks 'qa2' and 'qa3' (Fig.~\ref{fig:rag_qa2_10}). This lack of success is attributable to the specific demands of these tasks, which require the retrieval of multiple (two or three) supporting facts to generate accurate responses. For example, in instances where the key facts are "Mary got the milk there." and "Mary travelled to the hallway.", with the query being "Where is the milk?", the retrieval system may successfully identify the first fact but fail to retrieve the second due to insufficient similarity between the question and the latter fact.  

\section{Recurrent memory transformer with retrieval}

\begin{figure*}[t]
\centering
{\fontfamily{cmss}\selectfont\footnotesize  a \hspace{10pt}  Recurrent Memory Transformer (RMT) \hspace{40pt}    b \hspace{50pt} RMT with retrieval (RMT-R) \hspace{80pt}  \hfill \\ 
\includegraphics[width=0.37\textwidth]{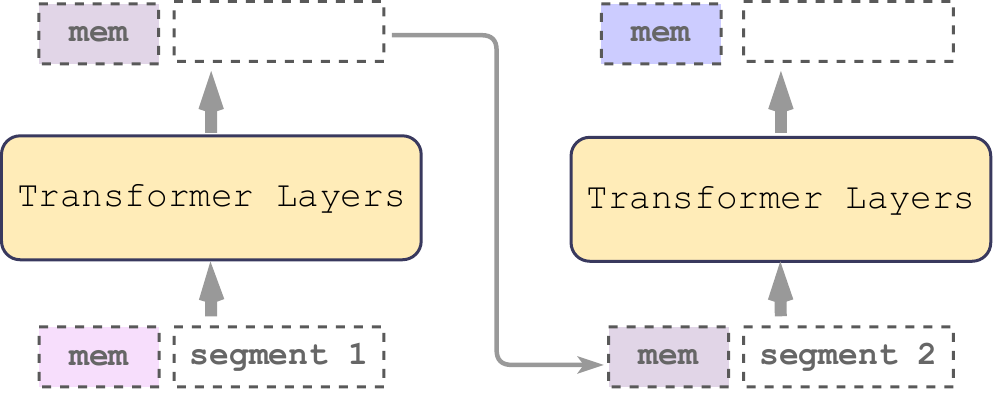} 
\hspace{30pt}
  \includegraphics[width=0.5\textwidth]{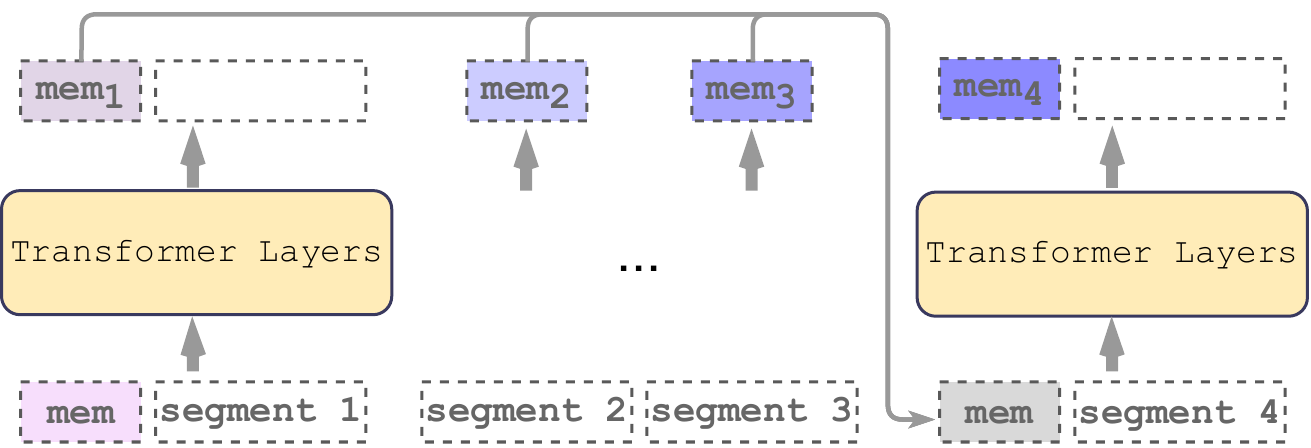}}
\caption{\textbf{Recurrent Memory Transformer with self-retrieval from memory.} (a) Recurrent memory transformer encodes information from the current segment to the memory vectors \texttt{[mem]}. Memory vectors from the previous segment are passed over to the next segment and updated. (b) With self-retrieval it processes each segment sequentially and collects the corresponding memory states. Here, while processing segment 4, the model can retrieve from previous states from segments 1-3. This overcomes the memory bottleneck that can occur in purely recurrent models, similar to attention in RNNs.}
\label{RMT_retrieve}
\end{figure*}

Recurrent models pass recurrent states between subsequent steps. Usually, recurrent states have a fixed size. At the end of sequence processing, a single recurrent state must hold information about the entire sequence, creating a bottleneck. To overcome this bottleneck, we can keep all previous recurrent states and allow the model to retrieve relevant previous states for the current processing step. In terms of RMT as an RNN, we propose incorporation of a self-retrieval (i.e., retrieval from self past states), which is conceptually equivalent to attention for RNNs~\citep{bahdanau2015neural}.

The Recurrent Memory Transformer (RMT)~\citep{rmt_2022} (see Fig.~\ref{RMT_retrieve}a) is an augmentation for Transformer models that extends their context size by segmenting sequences and processing them recurrently, resulting in linear scaling with input size. Memory, implemented as special memory tokens, is processed alongside segment tokens to enable the Transformer to access information from previous segments.

At each time step $t$, RMT processes an input segment $X^t$ along with memory state $M^{t-1}$ passed from the previous time step. RMT outputs updated memory state $M^t$ and model predictions $O^t$:
\begin{equation}
    O^t, M^t = \mathrm{RMT}(X^t, M^{t-1})\,.
\end{equation}

We implement self-retrieval for RMT (see Fig.~\ref{RMT_retrieve}b) by using cross-attention between all past states $[M^0, \dots, M^{t-1}]$ and previous memory state $M^{t-1}$. We follow \citep{vaswani2017attention}, but we use single-head attention:
\begin{equation*}
\begin{gathered}
    Q^t = M^{t-1} W^Q\,,\\
    K^t, V^t = [M^0, \dots, M^{t-1}] W^K, [M^0, \dots, M^{t-1}] W^V\,,\\
    R^t = \mathrm{Attention(Q^t, K^t, V^t)} W^O\,,
\end{gathered}
\end{equation*}
here $R^t$ are retrieved memories from the past states. Next, we augment RMT with retrieved states by concatenating the memory state from the previous step $M^{t-1}$ and $R^t$:
\begin{equation}
    O^t, M^t = \mathrm{RMT}(X^t, [M^{t-1}, R^t])\,.
\end{equation}

To account for the causal attention mask in decoder-only transformer models read memory tokens are placed at the beginning of the sequence and write memory tokens at the end $[M^{t-1}, X^t, M^{t-1}]$ as in \cite{rmt_2022}. The output from the write memory tokens is used as the updated memory state. Adding two types of memory is not required for encoder-only models. For RMT with self-retrieval (RMT-R), we concatenate retrieved states $R^t$ to read memory only, feeding $[M^{t-1}, R^t, X^t, M^{t-1}]$ as an input for RMT-R backbone transformer.

In RMT, each memory state $M^t$ has shape $m \times d$ and consists of $m$ memory tokens of size $d$. Thus, each memory token in memory state $M^t$ independently retrieves relevant memory tokens from a set of memory tokens from all previous memory states. Multiple memory tokens can represent different views of past inputs, like multiple heads in Transformer's attention.

Compared to RMT, RMT-R retains all previous memory states, therefore required storage to scale linearly with a number of segments $n$: $n\times m \times d$. Storing hidden states for all previous tokens would significantly increase this value as $m \ll L$, where $L$ is a segment length. There is also a small computational overhead to compute cross-attention for retrieval, which also scales linearly with the input length as $\mathcal{O}((1 \times m) \times (n \times m))$. In practice, we could infer RMT-R on up to 10M token sequence lengths and have not encountered a limit.

\section{RMT and RMT-R on BABILong}

\begin{figure*}[t]
\centering
\includegraphics[width=\textwidth]{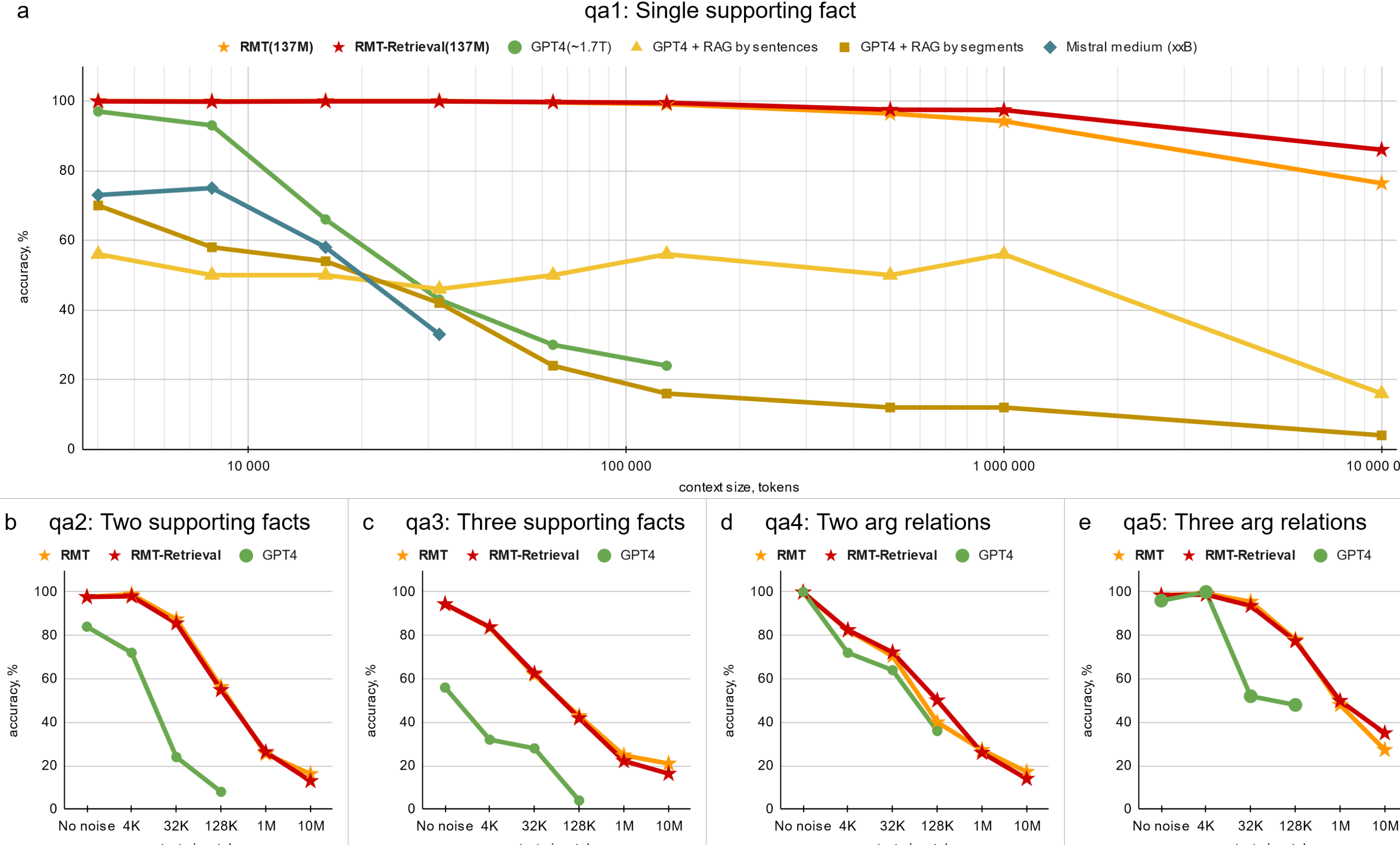}

\caption{\textbf{RMT and RMT-R significantly outperform LLMs with and without retrieval on extremely long needle in a haystack tasks.} Every task consists of a question related to one or few facts distributed inside textual context. The answer requires (a) one, (b) two or (c) three supporting facts amongst a set of irrelevant facts; (d) two or (e) three argument relations over multiple statements. An experiment with no distractor text added to the input is labeled as 'No noise'. }
\label{fig:RMT_10M}
\end{figure*}

RMT and RMT-R with a GPT-2~\citep{radford2019GPT2} backbone model are trained on each task individually with a segment size of 512 and memory size of 16. Train and evaluation splits of each task contain 10000 and 1000 samples respectively with a number of facts in each sample ranging from 2 to 320 depending on the task. A curriculum strategy is employed, initiating training on short sequences that fit in a single segment and then gradually introducing longer sequences once the training converges. During each curriculum stage $n$, sequence lengths are randomly sampled from $1$ to $n$ segments. We provide details of training and evaluation procedures in Appendix~\ref{appendix:rmt_training}.

Models trained on 32 segments demonstrate strong performance on sequence lengths up to 16k tokens present in the training set. Notably, both RMT versions outperform GPT-4 significantly, underscoring the efficiency of memory mechanism. 
Even more importantly, the superiority of recurrent models extends on out-of-domain sequences longer than the training size. By leveraging their emergent generalization abilities, both RMT and RMT-R exhibit consistent performance on longer sequences, reaching 128k tokens, with only a marginal quality degradation. The enhanced capabilities of RMT-R become particularly evident due to the additional information from retrieved memory states. 

Surprisingly, even with context sizes scaling to 1 million and even 10 million tokens, which is over 600 times of the training length, recurrent models persistently outperform their larger counterparts utilizing RAG. This remarkable leap in reasoning quality across extremely long sequences underscores the potential of recurrence paired with trainable self-retrieval mechanism.

\begin{figure*}[t]
\centering
\includegraphics[width=\textwidth]{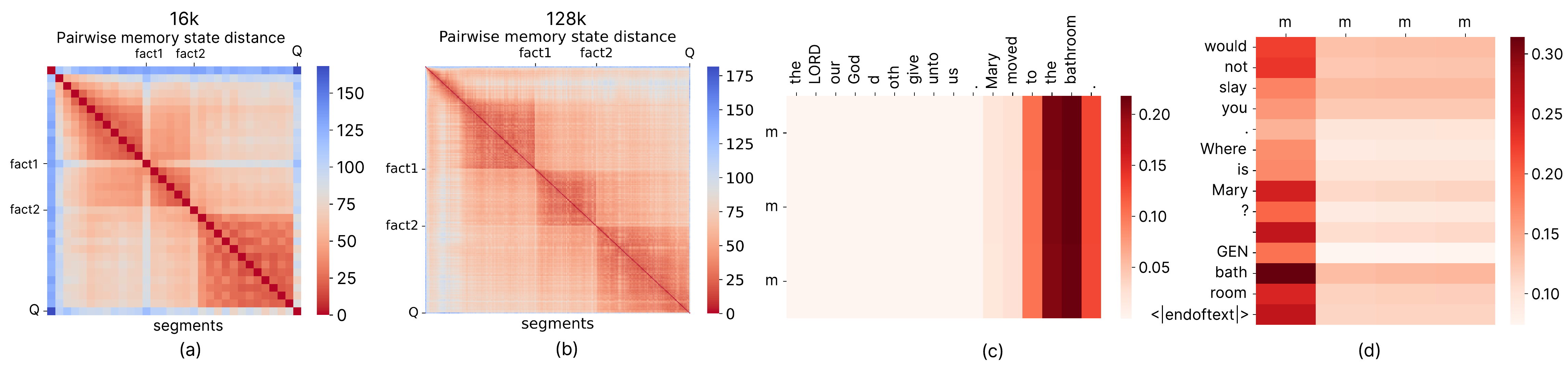}
\caption{\textbf{RMT learns to detect and store relevant facts using memory.} Heatmaps (a) and (b) represent pairwise distances between memory states on qa1 with context size 16k (a) and 128k (b). Distant states are marked with blue color and similar ones with red. Changes in memory mainly occurs when the model meets a new fact, which indicates model adaptation to distinguishing and storing facts in memory. Memory attention maps (c) and (d) when RMT writes the fact to memory (c) and then reads it when answering the question (d). The intensity of red color corresponds to the amount of attention between the query on the left and key on the top.}
\label{fig:RMT_analysis}
\end{figure*}

To understand how recurrent models consistently retain their performance over extremely long sequences, we analyze the RMT memory states and attention patterns on the qa1 task. We evaluate RMT trained on 32 segments or approximately 16k tokens on a single sample with two facts, see Figure~\ref{fig:RMT_analysis} (a) and (b). For both sequence lengths 16k and 128k the memory states exhibit a consistent pattern. In the absence of fact in input, the memory remains similar to its initial states, but the introduction of fact leads to visible change in the memory state. This indicates that the model learned to distinguish important facts from the background text and preserving them in memory until a question appears. The operations with memory are represented by distinctive patterns on attention maps, specifically, the process of writing a new fact to memory Figure~\ref{fig:RMT_analysis} (c) and reading from memory to answer a question (d). This visual demonstration supports the intuition of learning distinct memory operations when dealing with information scattered across extended contextual spans.  

\section{Related Work}

A new set of datasets~\cite{bai2023longbench, l_eval_an2023} and benchmarks~\cite{zeroscrolls_shaham2023} specifically designed to test the ability of LLMs to handle long contexts has been proposed.  The LongBench dataset~\cite{bai2023longbench} contains 6 types of real and synthetic problems, ranging from summarisation and multidoc QA to code completion. The average sample lengths in LongBench are 6k and 13k tokens for English and Chinese respectively. ZeroSCROLLS~\cite{zeroscrolls_shaham2023} focuses on testing models in few-shot learning with long context and consists only of validation and tasting data. L-Eval~\cite{l_eval_an2023} mostly combines smaller long sequence datasets and adds 4 newly annotated tasks. The average length of examples for L-Eval varies from 3 to 60 thousand tokens. Long Range Arena (LRA)~\cite{tay2021long} is a set of tasks with lengths from 1 to 16 thousand tokens. However, it mainly consists of very specific tasks such as ListOps (2k tokens), Byte-Level Text Classification (4k tokens)  and Byte-Level Text Retrieval (8k tokens), and others that are less related to NLP. They are not well suited for evaluating of modern LLMs without fine-tuning on these tasks and cannot fully represent the capabilites of GPT-4 with supported 100k+ tokens.
 
 All these datasets present a lot of different and challenging tasks however they are limited in length of examples. In this work, we propose a new approach to build datasets and to test abilities of language models to find and reason about specific information within a context of mullions 
 of tokens. We employ this approach for bAbI tasks resulting in bAbILong benchmark with infinitely long contexts.
 
In retrieval augmented generation (RAG), a language model is combined with a separate module, called a retriever. Given a specific request, the retriever finds a set of relevant parts from a dedicated data storage.
Then parts selected by the retriever along with the input are incorporated by the language model to make predictions. Many different implementations of the retrieval mechanism have been proposed~\cite{realm_guu2020, retro_borgeaud2022, replug_shi2023}. Some works focus on directly retrieving predictions~\cite{knn_lm_khandelwal2019}. Other works retrieve individual input tokens or text segments and add them to the LM input~\cite{realm_guu2020, retro_borgeaud2022}. For example, in REALM~\cite{realm_guu2020} whole text segments are retrieved and appended to the input to improve masked language modeling. In Memorizing Transformer~\cite{wu2022memorizing}, the retriever returns cached (key, value) pairs saved from previous training steps of the language model.
In Retrieval-Pretrained Transformer~\citep{rubin2023long}, an LM component processes long documents in chunks and a retriever finds relevant chunks. Representations of retrieved chunks are fused with current chunk in the LM component, and both the LM and retrieval parts are trained jointly.
AutoCompressor~\cite{chevalier-etal-2023-adapting} combines RMT-like~\citep{rmt_2022} approach with retrieval from external corpora. AutoCompressor is first used to produce memory tokens (or summary vectors) for text chunks. Next, off-the-shelf retriever is used and corresponding chunk's memory tokens are added to the context of the model.

In this work, we augment the Recurrent Memory Transformer~\cite{bulatov2023scaling} with the ability to retrieve its own past memory tokens. As far as we know, this is the first combination of a recurrent transformer with a trainable retrieval mechanism.

Recurrence is another mechanism to deal with long context~\cite{graves2014neural, legendre_voelker2019, memup_sorokin2022}.  Instead of processing the entire context, a recurrent model breaks it down into smaller segments. The recurrent hidden state acts as an aggregator of information from past segments of the sequence. Attending to a memory state is much cheaper than to all contexts. Many different architectures adding recurrence to transformers have been proposed~\cite{wu2020memformer, lei2020mart, fan2020feedback-transformer}.
For example, Compressive Transformer~\cite{rae2019compressive} updates recurrent memory by compressing hidden activation's from the previous segment to a smaller set of representations.
Recurrent Memory Transformer~\cite{rmt_2022} recurrently passes states of special memory tokens added to the input of Transformer.
Activation Beacon~\cite{zhang2024soaring} compresses activations from prior segments using separate parameters and integrates a sliding window mechanism, handling up to 400k tokens.
Temporal Latent Bottleneck~\cite{tlb_didolkar2022} Transformer splits computation into two streams: recurrent slow stream and fast stream with self-attention between tokens.  Block-Recurrent Transformer~\cite{hutchins2022blockrecurrent} employs LSTM-style~\cite{lstm_1999hochreiter} gates to update its recurrent state.

We use RMT in our experiments because of its simplicity, plug-and-play compatibility with pre-trained transformer-based language models, and promising scaling capabilities~\cite{bulatov2023scaling}.

Big Bird~\cite{zaheer2020big}, Longformer~\cite{beltagy2020longformer}, LongNet~\cite{ding2023longnet} help extend context length for Transformers by switching from full self-attention to sparse self-attention mechanisms with linear complexity. Works like RWKV~\cite{peng2023rwkv}, S4~\cite{gu2021s4}, Mamba~\cite{gu2023mamba}, take another approach and focus on advancing recurrent networks to reach high parallelism levels available to Transformers while retaining the linear complexity of RNN. These works show promising results on long sequences but are still lagging behind the best transformer models in natural language processing tasks. Mamba, however, seeks to bridge this gap.

\section*{Conclusions}

This work explores capability of current transformer based solutions in extremely long context processing. For that we propose BABILong, a novel benchmark designed to assess the performance of NLP models on long documents with distributed facts. The benchmark offers algorithmically adaptable document length and content placement, making it leak-proof for future large language models (LLMs). 

Our findings reveal limitations in popular LLMs like GPT-4 and RAG regarding effective long context utilization. Their performance heavily relies on the first 25 \% of the input, highlighting the need for improved context processing mechanisms.

We demonstrate the effectiveness of recurrent memory augmentation of transformer models. This approach achieves superior performance on documents up to 11 million tokens. Notably, the recurrent memory enables multi-hop reasoning, a crucial ability for tackling complex tasks. Attention map analysis provides valuable insights into how the model leverages the stored memory to identify relevant facts. 

This work not only sets a new record for processing of 11 million tokens but also strengthens the hypothesis that recurrent memory excels at filtering out irrelevant information compared to solely relying on attention mechanisms.

While fine-tuning required for memory augmentation, the use of a smaller GPT-2 model suggests significant potential for further improvement. We hypothesize that applying both the recurrent memory and retrieval to larger models can unlock even better generalization to longer contexts, both during training and inference. Exploring this avenue remains an exciting direction for future research.

\section*{Limitations}
The BABILong benchmark uses background texts to hide facts in them. In our experiments, we only tried PG19 and Wiki as background text sources. Other background texts may have a different effect on the results. PG19 and Wiki were chosen because they contain natural narratives and facts about people, in a way similar to bAbI tasks. Interference between similar facts in the background text can make the benchmark even more difficult. 

In GPT-4 and RAG experiments, we do not optimize the retrieval component and use only a popular one from OpenAI. We tried several prompts for GPT-4 and Mistral models, but the ones that we selected could be suboptimal. We provide them in Appendix~\ref{sec:prompt}.

Compared to RMT, RMT-R can reach the memory limit when predicting on very long sequences. RMT-R collects all past memory states, which leads to a linear space complexity with the number of segments. However, we were able to run it on sequences up to 10M tokens and did not hit the limit on A100 80Gb.

Although recurrent approaches, like RMT,  are hindered by their sequential nature, resulting in reduced parallelizability, they compensate by constant memory requirements, but it is also their limitation as storage capacity of the model is finite.




\bibliography{icml24}
\bibliographystyle{icml2024}

\newpage
\appendix
\onecolumn
\section{RMT and RMT-R training and evaluation details}
\label{appendix:rmt_training}
For RMT and RMT-R we use curriculum training with sequentially increasing number of segments. RMT uses the following schedule for number of segments: 1-2-4-6-8-16-32, RMT-R schedule: 2-4-6-8-16-32. We skip the single segment training for RMT-R, as it is equivalent to RMT in this case. During each curriculum stage $n$ the number of segment is chosen randomly from $1$ to $n$. We select learning rate from \{5e-05, 3e-05\} and use the AdamW optimizer and linear learning rate scheduling with warmup. We use a total batch size of $64$ and train for \{5000, 10000\} steps with early stopping if metrics stop increasing. For the backbone transformer we use the pretrained GPT-2 137M from HuggingFace: \url{https://huggingface.co/GPT-2}. We used up to 4 Nvidia A100 80Gb per experiment.

We evaluate models on the BABILong benchmark. We use full test set for sequences up to 1M tokens, for 10M tokens we only provide results for the first 100 samples. As shown in Table~\ref{tab:evaluation_time}, evaluation time grows linearly with context length. We fix a random seed used to sample background texts for the test set. However, the seed was not fixed to avoid overfitting to specific sampled texts during training.

\begin{table}[h]
\caption{\small Time required for processing 1000 BABILong samples with RMT using a single A100 80Gb GPU, including input data processing.}
\label{tab:evaluation_time}
\begin{center}
\begin{small}
\begin{sc}
\fontsize{7}{8}
\selectfont 
\begin{tabular}{lcccc}
\toprule
Context size                & 4k & 32k  & 128k & 1M  \\
\midrule
Processing time, minutes    & 4  & 30   & 80   & 315 \\

\bottomrule
\end{tabular}
\end{sc}
\end{small}
\end{center}
\end{table}

\section{Prompts used to benchmark GPT-4-Turbo and Mistral models}\label{sec:prompt}

We used the same prompts to evaluate GPT-4-Turbo and Mistral models in our tasks. Each prompt starts with the description of the task followed by several examples inside the $<$example$>$ $<$/example$>$ tags. The next section inside $<$context$>$ $<$/context$>$ tags contains an instance of the task. We additionally duplicate the question with the QUESTION mark, in order for the model recognize the question in the large input prompts. The last sentences specify the required response format. 

\begin{tcolorbox}[title=qa1 task]
\begin{Verbatim}[fontsize=\small]
I will give you context with the facts about positions of different persons hidden in some random text and a question. You need to answer the question based only on the information from the facts. If a person was in different locations, use the latest location to answer the question.

<example>
Charlie went to the hallway. Judith come back to the kitchen. Charlie travelled to balcony. Where is Charlie?
Answer: The most recent location of Charlie is balcony.
</example>

<example>
Alan moved to the garage. Charlie went to the beach. Alan went to the shop. Rouse travelled to balcony. Where is Alan?
Answer: The most recent location of Alan is shop.
</example>

<context>
{qa1 query with noise}
</context>

QUESTION: {qa1 question}

Always return your answer in the following format: The most recent location of 'person' is 'location'. Do not write anything else after that.
\end{Verbatim}
\end{tcolorbox}

\begin{tcolorbox}[title=qa2 task]
\begin{Verbatim}[fontsize=\small]
I give you context with the facts about locations and actions of different persons hidden in some random text and a question. You need to answer the question based only on the information from the facts. 

If a person got an item in the first location and travelled to the second location the item is also in the second location.
If a person dropped an item in the first location and moved to the second location the item remains in the first location.

<example>
Charlie went to the kitchen. Charlie got a bottle. Charlie moved to the balcony. Where is the bottle?
Answer: The bottle is in the balcony.
</example>

<example>
Alan moved to the garage. Alan got a screw driver. Alan moved to the kitchen. Where is the screw driver?
Answer: The screw driver is in the kitchen.
</example>

<context>
{qa2 query with noise}
</context>

QUESTION: {qa2 question}

Always return you answer in the following format: The 'item' is in 'location'. Do not write anything else after that.
\end{Verbatim}
\end{tcolorbox}

\begin{tcolorbox}[title=qa3 task]
\begin{Verbatim}[fontsize=\small]
I give you context with the facts about locations and actions of different persons hidden in some random text and a question.
You need to answer the question based only on the information from the facts.

If a person got an item in the first location and travelled to the second location the item is also in the second location.
If a person dropped an item in the first location and moved to the second location the item remains in the first location

<example>
John journeyed to the bedroom.Mary grabbed the apple. Mary went back to the bathroom. Daniel journeyed to the bedroom. Daniel moved to the garden. Mary travelled to the kitchen. Where was the apple before the kitchen?
Answer: Before the kitchen the apple was in the bathroom.
</example>

<example>
John went back to the bedroom. John went back to the garden. John went back to the kitchen. Sandra took the football. Sandra travelled to the garden. Sandra journeyed to the bedroom. Where was the football before the bedroom?
Answer: Before the kitchen the football was in the garden.
</example>
                    
<context>
{qa3 query with noise}
</context>
                    
QUESTION: {qa3 question}

Always return you answer in the following format: Before the $location_1& the $item$ was in the $location_2$. Do not write anything else after that.
\end{Verbatim}
\end{tcolorbox}

\begin{tcolorbox}[title=qa4 task]
\begin{Verbatim}[fontsize=\small]
I will give you context with the facts about different people, their location and actions, hidden in some random text and a question.
You need to answer the question based only on the information from the facts.

<example>
The hallway is south of the kitchen. The bedroom is north of the kitchen. What is the kitchen south of?
Answer: bedroom              
</example>

<example>
The garden is west of the bedroom. The bedroom is west of the kitchen. What is west of the bedroom?
Answer: garden
</example>

<context>
{qa4 query with noise}
</context>

QUESTION: {qa4 question}

Your answer should contain only one word - location. Do not write anything else after that
\end{Verbatim}
\end{tcolorbox}

\begin{tcolorbox}[title=qa5 task]
\begin{Verbatim}[fontsize=\small]
I will give you context with the facts about locations and their relations hidden in some random text and a question. You need to answer the question based only on the information from the facts.

<example>
Mary picked up the apple there. Mary gave the apple to Fred. Mary moved to the bedroom. Bill took the milk there. Who did Mary give the apple to?
Answer: Fred        
</example>

<example>
1 Jeff took the football there. Jeff passed the football to Fred. Jeff got the milk there. Bill travelled to the bedroom. Who gave the football?
Answer: Jeff
</example>

<example>
Fred picked up the apple there. Fred handed the apple to Bill. Bill journeyed to the bedroom. Jeff went back to the garden. What did Fred give to Bill?
Answer: apple
</example>

<context>
{qa5 query with noise}
</context>
                        
QUESTION: {qa5 question}

Your answer should contain only one word. Do not write anything else after that. Do not explain your answer.
\end{Verbatim}
\end{tcolorbox}

\section{Analysis of LLM performance for different locations of the supporting facts}\label{sec:llmloc}

\begin{figure}[t]
\centering
\includegraphics[width=0.5\linewidth]{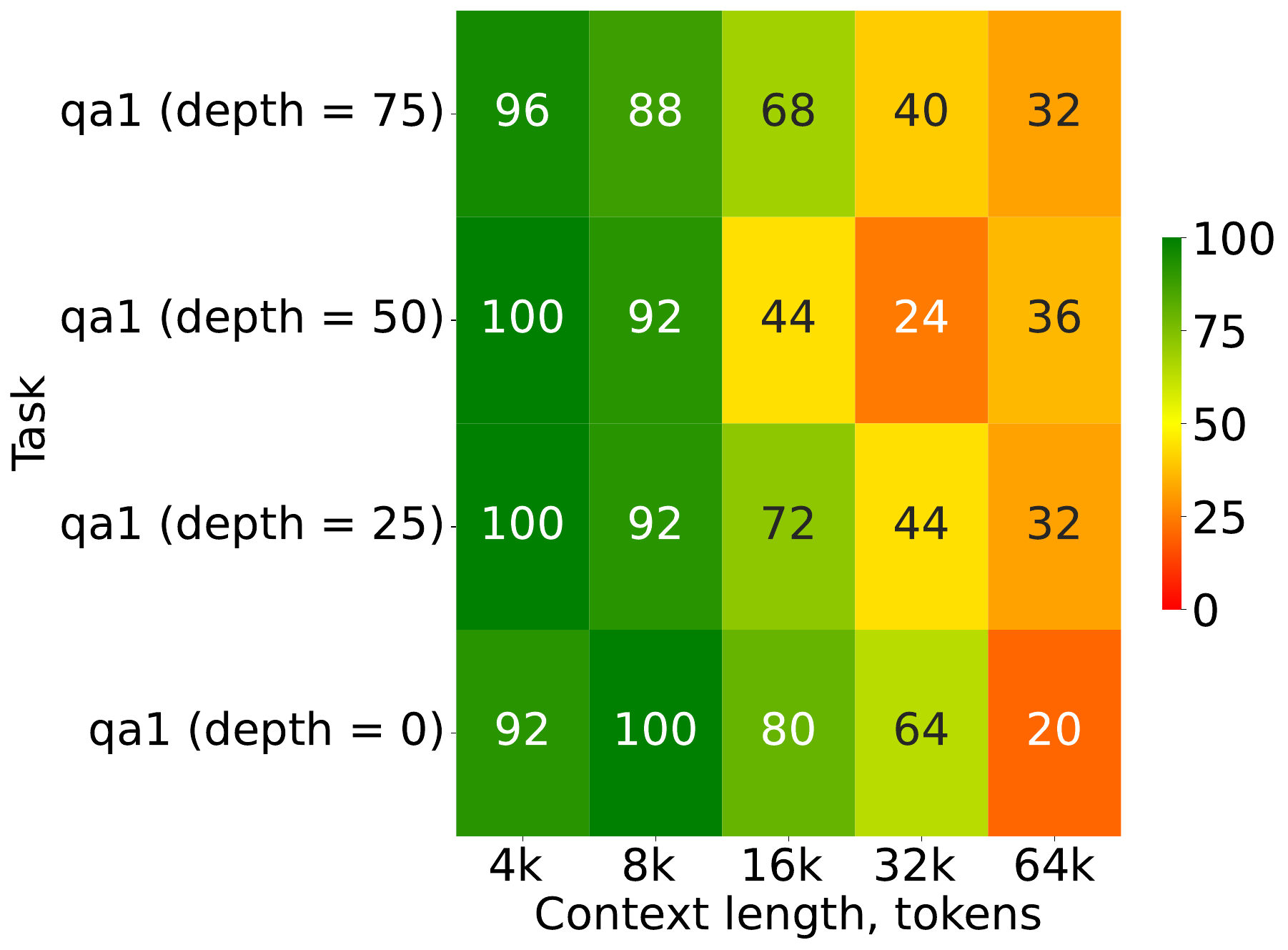}
\caption{Evaluation results for GPT-4-Turbo with different locations of the facts in the qa1 task.}
\label{fig:gpt_full}
\end{figure}

Fig.~\ref{fig:gpt_full} shows the evaluation result of the GPT-4-Turbo model when all the facts in task are located in the same quarter of the input query. It is seen that the performance of the model is not the same for different locations of the supporting facts. The most difficult location to identify the facts is in the middle of context which corresponds to the depts = 50 in the Fig.~\ref{fig:gpt_full}.

\newpage

\section{Needle in a Haystack Task Examples}
\label{appendix:babi_samples}
This section contains samples of the final BABILong dataset for first five tasks. First part of each example displays facts needed to solve the task, second part shows the example with background text with total length up to 512 tokens, and the final part contains question and the desired answer. The tasks differ by the number of facts and the task complexity, testing the ability for multiple reasoning aspects.

\begin{tabularx}{\textwidth}{X}
  \textbf{qa1 single-supporting-fact}\\
\toprule 

\textbf{Facts:} Sandra moved to the kitchen. Sandra went back to the garden. Sandra journeyed to the office. Mary moved to the office. Sandra journeyed to the bathroom. Daniel moved to the office. Daniel went back to the kitchen. Mary moved to the hallway.
\\
\midrule
\textbf{Input:} Now this loss of the sense of proportion in human affairs, Sir, is a very bad sign, and a well-nigh infallible indicator of nerve-strain and general overpressure. \textbf{Sandra moved to the kitchen.} But I find a yet more unmistakable evidence in support of my contention in the extraordinary emotional sensibility revealed by these headlines whenever some unfortunate person has been sentenced to death for the most commonplace murder. There is clearly a profound conviction that the jury who heard the evidence, the judge who pronounced their verdict of guilty, the only possible conclusion they could reasonable come to, and the HOME SECRETARY who found himself unable to recommend a reprieve, were, one and all, engaged in a cold-blooded conspiracy against a perfectly innocent man. The convict has said to himself, and that seems to be considered sufficient. And so, night after night, the authors of these headlines harrow themselves by announcing such items as "Blank protests his innocence to his Solicitor." "Distressing Scene on the Scaffold." \textbf{Sandra went back to the garden.} Consider the strain of all these alterations of hope and despair, repeated time after time, and almost invariably without even the consolation of deferring the fate of their protege by a single hour! \textbf{Sandra journeyed to the office.} Is it not too much for the strongest constitution to endure? a service which the society has no right to demand from any of its members? Yes, Sir, whether these devoted servants of the public know it or not, they are running a most frightful risk; the word which hangs above their heads may fall at any moment. \textbf{Mary moved to the office.} \textbf{Sandra journeyed to the bathroom.} \textbf{Daniel moved to the office.} Suppose, for example--and it is surely not wholly an imaginary danger I foresee--suppose that some day some event should happen somewhere of real and serious importance. \textbf{Daniel went back to the kitchen.} \textbf{Mary moved to the hallway.} Have they left themselves any epithet in reserve capable of expressing their sensations at all adequately? They have not; they have squandered participles 
and adjectives in such reckless profusion that they will discover they are reduced to the condition of inarticulate bankrupts; and, speaking as a medical man,  ...
\\
\midrule
\textbf{Question:} Where is Mary?
\textbf{Answer:} hallway
\\
\bottomrule
\end{tabularx}

\begin{tabularx}{\textwidth}{X}
  \textbf{qa2 two-supporting-facts}\\
\toprule 
\textbf{Facts:} John journeyed to the garden. John grabbed the apple there. Mary travelled to the hallway. Mary went back to the bathroom. Mary went to the garden. Mary travelled to the office. Daniel went to the office. Daniel went to the bedroom. Sandra went back to the office. Sandra journeyed to the garden. Mary travelled to the kitchen. Daniel moved to the kitchen. John put down the apple. Daniel journeyed to the garden. Sandra went to the bathroom. John got the apple there. Daniel travelled to the bedroom. Sandra moved to the hallway. John discarded the apple. Mary travelled to the garden.
\\
\midrule
\textbf{Context:} "From what I have already observed," said Mr. \textbf{John journeyed to the garden.} \textbf{John grabbed the apple there.} \textbf{Mary travelled to the hallway.} \textbf{Mary went back to the bathroom.} Ellison, "you will understand that I reject the idea, here expressed, of 'recalling the original beauty of the country.' \textbf{Mary went to the garden.} The original beauty is never so great as that which may be introduced. Of course, much depends upon the selection of a spot with capabilities. What is said in respect to the 'detecting and bringing into practice those nice relations of size, proportion and color,' is a mere vagueness of speech, which may mean much, or little, or nothing, and which guides in no degree. \textbf{Mary travelled to the office.} \textbf{Daniel went to the office.} \textbf{Daniel went to the bedroom.} That the true 'result of the natural style of gardening is seen rather in the absence of all defects and incongruities, than in the creation of any special wonders or miracles,' is a proposition better suited to the grovelling apprehension of the herd, than to the fervid dreams of the man of genius. \textbf{Sandra went back to the office.} \textbf{Sandra journeyed to the garden.} The merit suggested is, at best, negative, and appertains to that hobbling criticism which, in letters, would elevate Addison into apotheosis. \textbf{Mary travelled to the kitchen.} In truth, while that merit which consists in the mere avoiding demerit, appeals directly to the understanding, and can thus be foreshadowed in Rule, the loftier merit, which breathes and flames in invention or creation, can be apprehended solely in its results. \textbf{Daniel moved to the kitchen.} \textbf{John put down the apple.} \textbf{Daniel journeyed to the garden.} Rule applies but to the excellences of avoidance--to the virtues which deny or refrain. \textbf{Sandra went to the bathroom.} \textbf{John got the apple there.} \textbf{Daniel travelled to the bedroom.} We may be instructed to build an Odyssey, but it is in vain that we are told how to conceive a 'Tempest,' an 'Inferno,' a 'Prometheus Bound,' a 'Nightingale,' such as that of Keats, or the 'Sensitive Plant' of Shelley. \textbf{Sandra moved to the hallway.} \textbf{John discarded the apple.} But, the thing  ...\textbf{Mary travelled to the garden.} 
\\
\midrule
\textbf{Question:} Where is the apple? \textbf{Answer:} garden
\\
\bottomrule
\end{tabularx}

\begin{tabularx}{\textwidth}{X}
  \textbf{qa3 three-supporting-facts}\\
\toprule 
\textbf{Facts:} Sandra travelled to the office. Sandra picked up the football there. Sandra journeyed to the garden. Sandra journeyed to the bathroom.
\\
\midrule
\textbf{Context:} "From what I have already observed," said Mr. \textbf{Sandra travelled to the office.} Ellison, "you will understand that I reject the idea, here expressed, of 'recalling the original beauty of the country.' The original beauty is never so great as that which may be introduced. Of course, much depends upon the selection of a spot with capabilities. What is said in respect to the 'detecting and bringing into practice those nice relations of size, proportion and color,' is a mere vagueness of speech, which may mean much, or little, or nothing, and which guides in no degree. That the true 'result of the natural style of gardening is seen rather in the absence of all defects and incongruities, than in the creation of any special wonders or miracles,' is a proposition better suited to the grovelling apprehension of the herd, than to the fervid dreams of the man of genius. \textbf{Sandra picked up the football there.} The merit suggested is, at best, negative, and appertains to that hobbling criticism which, in letters, would elevate Addison into apotheosis. In truth, while that merit which consists in the mere avoiding demerit, appeals directly to the understanding, and can thus be foreshadowed in Rule, the loftier merit, which breathes and flames in invention or creation, can be apprehended solely in its results. Rule applies but to the excellences of avoidance--to the virtues which deny or refrain. \textbf{Sandra journeyed to the garden.} We may be instructed to build an Odyssey, but it is in vain that we are told how to conceive a 'Tempest,' an 'Inferno,' a 'Prometheus Bound,' a 'Nightingale,' such as that of Keats, or the 'Sensitive Plant' of Shelley. But, the thing done, the wonder accomplished, and the capacity for apprehension becomes universal. \textbf{Sandra journeyed to the bathroom.} The sophists of the negative school, who, through inability to create, have scoffed at creation, are now found the loudest in applause. What, in its chrysalis condition of principle, affronted their demure reason, never fails, in its maturity of accomplishment, to extort admiration from their instinct of the beautiful or of the sublime. "  ...
\\
\midrule
\textbf{Question:} Where was the football before the bathroom? \textbf{Answer:} garden
\\
\bottomrule
\end{tabularx}

\begin{tabularx}{\textwidth}{X}
  \textbf{qa4 two-arg-relations}\\
\toprule 
\textbf{Facts:} The garden is south of the bathroom. The bedroom is north of the bathroom.
\\
\midrule
\textbf{Context:} 'A mixture of pure art in a garden scene, adds to it a great beauty.' This is just; and the reference to the sense of human interest is equally so. I repeat that the principle here expressed, is incontrovertible; but there may be something even beyond it. There may be an object in full keeping with the principle suggested--an object unattainable by the means ordinarily in possession of mankind, yet which, if attained, would lend a charm to the landscape-garden immeasurably surpassing that which a merely human interest could bestow. \textbf{The garden is south of the bathroom.} The true poet possessed of very unusual pecuniary resources, might possibly, while retaining the necessary idea of art or interest or culture, so imbue his designs at once with extent and novelty of Beauty, as to convey the sentiment of spiritual interference. It will be seen that, in bringing about such result, he secures all the advantages of interest or design, while relieving his work of all the harshness and technicality of Art. \textbf{The bedroom is north of the bathroom.} In the most rugged of wildernesses--in the most savage of the scenes of pure Nature--there is apparent the art of a Creator; yet is this art apparent only to reflection; in no respect has it the obvious force of a feeling. Now, if we imagine this sense of the Almighty Design to be harmonized in a measurable degree, if we suppose a landscape whose combined strangeness, vastness, definitiveness, and magnificence, shall inspire the idea of culture, or care, or superintendence, on the part of intelligences superior yet akin to humanity--then the sentiment of interest is preserved, while the Art is made to assume the air of an intermediate or secondary Nature--a Nature which is not God, nor an emanation of God, but which still is Nature, in the sense that it is the handiwork of the angels that hover between man and God." It was in devoting his gigantic wealth to the practical embodiment of a vision such as this--in the free exercise in the open air, which resulted from personal direction of his plans--in the continuous and unceasing object which these plans afford--in the contempt of ambition which it enabled him more to feel than to affect  ...
\\
\midrule
\textbf{Question:} What is south of the bathroom? \textbf{Answer:} garden
\\
\bottomrule
\end{tabularx}

\begin{tabularx}{\textwidth}{X}
  \textbf{qa5 three-arg-relations}\\
\toprule 
\textbf{Facts:} Fred grabbed the football there. Jeff took the apple there. Jeff dropped the apple. Bill picked up the apple there. Mary travelled to the kitchen. Mary went back to the hallway. Bill went to the garden. Fred travelled to the garden. Bill passed the apple to Fred. Fred left the apple. Fred went back to the hallway. Fred handed the football to Mary.
\\
\midrule
\textbf{Context:} It was evident that the besiegers were in no hurry; that they were living upon the provisions left in the valley; and that it was their intention to reduce the besieged by famine. \textbf{Fred grabbed the football there.} \textbf{Jeff took the apple there.} In fact the inhabitants of the Val d'Avon had been able to carry with them only a small quantity of provisions. \textbf{Jeff dropped the apple.} We have described the three kinds of porcelain made in Hizen for exportation to Europe, and we have seen that by the middle of the seventeenth century this commerce, in the hands of the Dutch, and to some extent of the Chinese, had already attained large proportions. Before turning to the kilns that sprung up in other parts of Japan during the eighteenth century--of these the origin in every case can be traced back directly or indirectly to the early Hizen factories--we must say a word about some other varieties of porcelain made in the same neighbourhood, but not destined for foreign use. \textbf{Bill picked up the apple there.} The village or town of Arita, of which the better-known Imari is the port, lies about fifty miles to the north-east of Nagasaki, and it may almost be regarded as the King-te-chen of Japan. \textbf{Mary travelled to the kitchen.} \textbf{Mary went back to the hallway.} The clay and china-stone used there is now brought, for the most part, from the adjacent islands, from Hirado, from Amakusa, and even from the more remote Goto islands. \textbf{Bill went to the garden.} By a combination of some of the most important potters of the district, and with the assistance of some wealthy merchants, a company, the Koransha, was formed some twenty-five years ago,[123] and an attempt was made to keep up the quality of the porcelain produced, at least from a technical point of view. \textbf{Fred travelled to the garden.} It was certainly time for some such effort to be made, for about that period, just after the Philadelphia Exhibition, the arts of Japan reached perhaps their nadir. \textbf{Bill passed the apple to Fred.} \textbf{Fred left the apple.} MIKÔCHI OR HIRADO WARE.--It was with a somewhat similar object that, ... \textbf{Fred went back to the hallway.} \textbf{Fred handed the football to Mary.} 
\\
\midrule
\textbf{Question:} What did Bill give to Fred? \textbf{Answer:} apple
\\
\bottomrule
\end{tabularx}

\end{document}